\pdfoutput=1
\documentclass[11pt]{article}

\usepackage[]{acl}

\usepackage{CJKutf8}
\usepackage{float}
\usepackage{graphicx}
\usepackage{makecell}
\usepackage{multirow}
\usepackage{threeparttable}
\usepackage{float}
\usepackage{graphicx}
\usepackage{makecell}
\usepackage{multirow}
\usepackage{booktabs}
\usepackage{threeparttable}
\usepackage{CJKutf8}

\usepackage{times}
\usepackage{latexsym}

\usepackage[T1]{fontenc}
\usepackage[utf8]{inputenc}
\usepackage{microtype}

\title{A Comprehensive Evaluation and Analysis Study for Chinese Spelling Check}

\author{Xunjian Yin \and Xiaojun Wan\\
  Wangxuan Institute of Computer Technology, Peking University \\
  Center for Data Science, Peking University \\
  The MOE Key Laboratory of Computational Linguistics, Peking University \\
  \texttt{\{xjyin, wanxiaojun\}@pku.edu.cn}}

\begin{document}
\begin{CJK*}{UTF8}{gkai}
\maketitle
\begin{abstract}
With the development of pre-trained models and the incorporation of phonetic and graphic information, neural models have achieved high scores in Chinese Spelling Check (CSC). 
However, it does not provide a comprehensive reflection of the models' capability due to the limited test sets. 
In this study, we abstract the representative model paradigm, implement it with nine structures and experiment them on comprehensive test sets we constructed with different purposes. We perform a detailed analysis of the results and find that: 1) Fusing phonetic and graphic information reasonably is effective for CSC. 
2) Models are sensitive to the error distribution of the test set, which reflects the shortcomings of models and reveals the direction we should work on. 3) Whether or not the errors and contexts have been seen has a significant impact on models. 
4) The commonly used benchmark, SIGHAN, can not reliably evaluate models' performance.
\end{abstract}

\section{Introduction}
Spelling errors are common in sentences not only written by people but also produced in natural language processing tasks, which are very harmful. 
Therefore, more and more methods have been proposed in the spelling check task  \citep{etoori2018automatic,guo2019spelling, zhang2020spelling}.

\begin{table*}[]
\centering
\resizebox{\textwidth}{!}{%
\begin{tabular}{c|ccccccccc}
\toprule
\textbf{Wrong}   & 语 & 言 & {\color[HTML]{FE0000} 适(pinyin: \emph{shi4}. "adjust")} & 有 & {\color[HTML]{FE0000} 现(strokes: \{王, 见\}. "now")} & 律 & 可 & 循 & 的 \\ \midrule
\textbf{Correct} & 语 & 言 & 是(pinyin: \emph{shi4}. "is")                        & 有 & 规(strokes: \{夫, 见\}. "rules")                        & 律 & 可 & 循 & 的 \\ \bottomrule
\end{tabular}%
}
\caption{Examples of phonological similarity error and visual similarity error. The correct sentence means "Language has rules to follow."}
\label{tab:example}
\end{table*}

Unlike English or other alphabetic languages, Chinese is based on characters, the number of which is more than 10K. 
Moreover, a large number of Chinese characters are similar either in phonology or in morphology so that they are easily to be misspelled into another character in the vocabulary and hard to be corrected  \citep{kukich1992techniques, jia2013graph, wang2019confusionset}. 
As illustrated in Table \ref{tab:example}, the original wrong sentence contains two incorrect characters in red: the first one is phonetic similarity error because both 适~and 是's pinyin\footnote{pinyin is the phonetic system of Mandarin Chinese.} are \emph{shi4}; the other is graphic similarity error between 现~and 规.
According to \citet{liu2011visually}, 83\% of Chinese spelling errors are caused by phonetic similarity, 48\% are due to graphic similarity and 35\% involve both factors. 

\begin{figure}
\centering 
\includegraphics[width=0.46\textwidth]{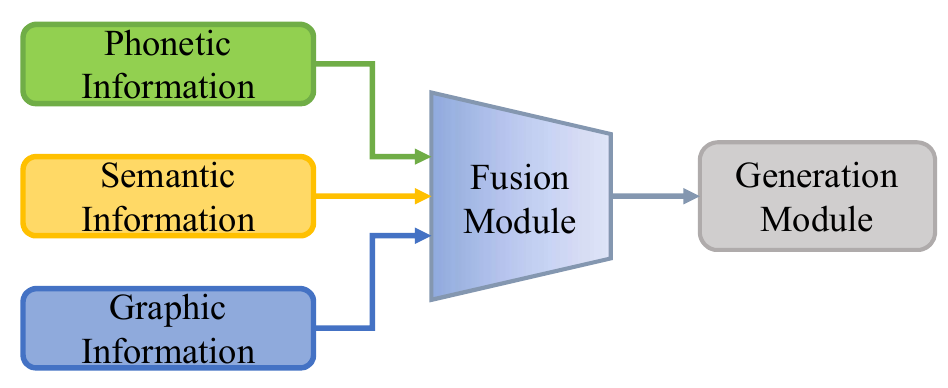} 
\caption{Abstraction of representative CSC models.} 
\label{Model}
\end{figure}

Therefore, in Chinese Spelling Check(CSC) task, a lot of work has been done trying to incorporate phonetic and graphic information into neural models to learn the phonology or morphology relationships between characters \citep{cheng2020spellgcn,nguyen2020domain,xu2021read, huang2021phmospell,wang2021dynamic, liu2021plome, ji2021spellbert, zhang2021correcting}. However, it is still not clear what the models have achieved and what techniques are really effective on the CSC task due to the limited test sets and lack of comprehensive evaluation and analysis. 

On the one hand, to facilitate our evaluation and  analysis, we abstract the structure of representative CSC models as a unified paradigm shown in Figure \ref{Model}.
In terms of phonetic information, features can be the encoding of pinyin sequences or the hidden representation of the pre-trained speech generation model such as Tacotron2 \citep{shen2018natural}.
As for the graphic information, it can be represented by the encoding of the strokes\footnote{\url{https://en.wikipedia.org/wiki/Stroke order}} sequences that compose characters or by the encoding of the font images from pre-trained models such as VGG \citep{simonyan2014very}.
And mainstream work always uses BERT \citep{devlin2018bert} to get semantic information.
As for the fusion module, its purpose is to converge these three types of information. 
Commonly used ways are to encode three types of information separately and then concatenate them together or to use a gating mechanism to fuse the information.
Finally, the generation module's objective can be corrected characters, maybe together with the corrected pinyin and strokes.

On the other hand, there are not enough benchmarks for CSC. The commonly used datasets are only the SIGHAN datasets \citep{wu2013chinese, yu2014overview, tseng2015introduction}. 
However, they are too small and lack in number and diversity of errors.
To compensate for these deficiencies and to more comprehensively evaluate the models' capabilities and performance in different dimensions, we designed multiple test sets for evaluation by carefully controlling the distribution of errors. 
For example, our test sets can reflect the models' performance facing sentences with different error frequencies and unknown errors, and the impact of the seen context or the seen errors, etc.

The contributions of this paper are summarized below:
1) We abstract the common CSC model paradigm and experimentally evaluate the different implementations.
2) We build several controlled test sets to fully evaluate and compare the models.
3) We obtain some useful and novel conclusions and advice from our experiments.
4) The code and datasets will be released to the community. 

Some representative conclusions and advice are listed here:
1) Fusing phonetic and graphic information reasonably is helpful. 
2) Models are sensitive to the error distribution of the test set.  We should pay more attention to it.
3) Whether or not the errors and contexts have been seen has a significant impact on the model. So we should consider the diversity of the confusion set and the domain of the text when performing data augmentation.
4) Character level metric is more stable and should be used to evaluate models. SIGHAN test sets can not reflect the model's performance reliably.

\section{Related Work}
CSC task has achieved great improvements in recent years.
FASpell \citep{hong2019faspell} applied BERT as a denoising autoencoder for CSC. Soft-Masked BERT \citep{zhang2020spelling} chose to combine a Bi-GRU based detection network and a BERT based correction network.

In recent times, many studies have attempted to introduce phonetic and graphic information into CSC models. SpellGCN was proposed to employ graph convolutional network on pronunciation and shape similarity graphs.
\citet{nguyen2020domain} employed TreeLSTM to get hierarchical character embeddings as graphic information. 
REALISE \citep{xu2021read} used Transformer \citep{vaswani2017attention} and ResNet5 \citep{he2016deep} to capture phonetic and graphic information separately.
In this respect，PLOME \citep{liu2021plome} chose to apply the GRU \citep{bahdanau2014neural} to encode pinyin and strokes sequence.
PHMOSpell \citep{huang2021phmospell} derived phonetic and graphic information from multi-modal pre-trained models including Tacotron2 and VGG19.

However, the benchmarks for CSC are very inadequate and little work has been done on the model evaluation. The widely used datasets are SIGHAN datasets \citep{wu2013chinese, yu2014overview, tseng2015introduction} which are used in CSC campaigns in 2013, 2014 and 2015. 

\citet{mita2021grammatical} evaluated the generalization capability of grammatical error correction models with controlled vocabularies. \citet{nagata2021exploring} explored the capacity of a large-scale masked language model to recognize grammatical errors. To our knowledge, no study has conducted a comprehensive review of CSC models.

\section{Models Construction}
As shown in Figure \ref{Model}, we abstract the representative CSC model paradigm. 
In order to more comprehensively evaluate and analyze the different model structures, we classify the models according to their sources of fused information and the way they fuse information.

\begin{figure*}
\centering 
\includegraphics[width=0.87\textwidth]{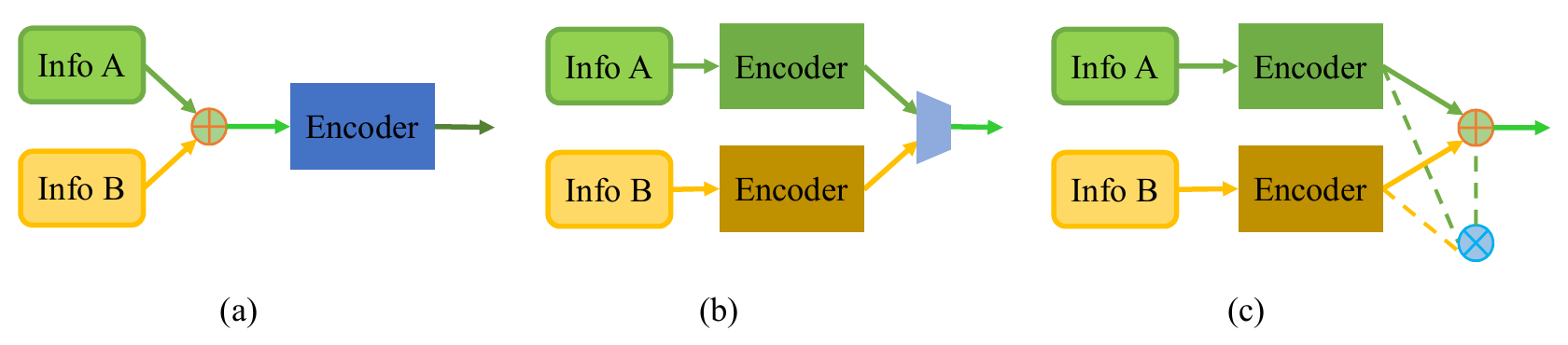} 
\caption{Illustration of information fusion methods. (a) is Add-Encode method, (b) is Encode-Transform method and (c) is Encode-Gate method.} 
\label{fuse}
\end{figure*}

\subsection{Information Source}
The input information can be divided into three categories: semantic, phonetic and graphic information.
According to the mainstream work, we use \textbf{BERT(BT)} to encode the characters in the sentence to capture the semantic information.
But for phonetic and graphic information, two different types of information sources are commonly used.

\textbf{Symbolic Sequences} \quad As shown in Table~\ref{tab:example}, phonetic information of a character can be represented by the pinyin sequence obtained by the character-phonics mapping Unihan Database\footnote{\url{http://www.unicode.org/charts/unihan.html}} and graphic information can be represented by the stroke sequence obtained via Chaizi Database\footnote{\url{https://github.com/kfcd/chaizi}}.
They can be noted as \textbf{PSym} and \textbf{GSym}.

\textbf{Multimodal Features} \quad Phonetic and graphic information of Chinese characters can also be obtained from speeches and images. 
Specifically, we derive them from intermediate representations of Tacotron2 \citep{shen2018natural} in text-to-speech task and VGG19 \citep{simonyan2014very} in computer vision task.
They can be noted as \textbf{PMod} and \textbf{GMod}.

\subsection{Information Fusion Method}
There are three common ways of fusing information as shown in Figure \ref{fuse}. The \emph{Encoder} inside is implemented by Transformer Encoder \citep{vaswani2017attention} for symbolic sequence information input. As for multimodal features, we think they have already encoded by the pre-trained model.

\textbf{Add-Encode(AE)} \quad Add different information into one feature vector directly and then encode it.

\textbf{Encode-Transform(ET)} \quad Encode different information separately and then transform them. The transform layer is a fully-connected layer. 

\textbf{Encode-Gate(EG)} \quad Fuse the different information through a gating mechanism. The gate values are computed by a fully-connected layer followed by a sigmoid function.

\begin{table*}[]
\centering
\resizebox{\textwidth}{!}{%
\begin{tabular}{c|c|c}
\toprule
\textbf{Information}               & \textbf{Model}      & \textbf{Introduction}                                                                           \\ \midrule
None                      & BERT       & Original BERT to encode the character sequence                                          \\ \midrule
\multirow{3}{*}{Phonetic} & BT-PSym-AE & Add the character and pinyin's embeddings  together and encode them   \\ 
                          & BT-PSym    & Pinyin sequences are encoded by Transformer and transformed with BERT's output \\ 
                          & BT-PMod    & Similar to BT-PSym, but the phonetic information is from Tacotron2                    \\ \midrule
\multirow{3}{*}{Graphic}  & BT-GSym-AE & Add the character and strokes' embeddings  together and encode them         \\ 
                          & BT-GSym   & Similar to BT-PSym, but the input information is stroke sequence                      \\ 
                          & BT-GMod    & Similar to BT-GSym, but the graphic information is from VGG19                          \\ \midrule
\multirow{2}{*}{Both}     & BT-PG      & Transform Tacotron2's phonetic, VGG19's graphic information and BERT's output \\ 
                          & BT-PG-EG    & Similar to BT-PG, but use gate mechanism to fuse the different   information                 \\ \bottomrule
\end{tabular}%
}
\caption{Implemented models and their introduction. Except for BT-PSym-AE, BT-GSym-AE and BT-PT-EG, all the others take the Encode-Transform(ET) method to fuse the information. And the transform layer in ET is a fully-connected layer.}
\label{tab:imp}
\end{table*}

\subsection{Implemented Models}
Since there are such a variety of modules and methods mentioned above, we combine them and conduct experiments on representative structures as shown in Table \ref{tab:imp}. 
Note that for models that add both phonetic and graphic information, we can try any combination of phonetic and graphic information sources to find the best combination.
In fact, much previous work has been done in this direction to  achieve the best results \citep{huang2021phmospell,liu2021plome,xu2021read}. 
Our work wants to demonstrate the benefits of fusing two types of information, and without loss of generality, we only experiment with the model that combines the widely-used phonetic and graphic multimodal information as a representative of information combination models.
The aspects of the different models compared are shown in Figure~\ref{modelcomp} in Appendix~\ref{modelcompa}. And the implementation details of models are illustrated in Appendix~\ref{detailmodel}.

\section{Datasets Construction}
We crawled Chinese news articles as our source dataset which contains a total of 2050K raw sentences.  
Their average length is 46 characters.
Among them, 5K sentences are extracted to construct the validation set, 5K sentences to construct different types of test sets, and the rest to construct the training set.

% \subsection{Methodology}

The structure of confusion set $C$ is a dictionary $\{k_1:\{v_{11},\dots, v_{1n}\}, \dots \}$, where $k_i$ is a Chinese character and $v_{ij}$ is the $j$th error-prone candidate for $k_i$.
$k_i$ and $v_{ij}$ form a misspelling pair $(k_i, v_{ij})$, where $v_{ij} \in C[k_i]$ ("$C[x]$" means getting the values of key $x$ in the $C$).
To create datasets, for each character $x$ in the raw sentence, we will replace it randomly with one of its candidate characters $x' \in C[x]$ according to the confusion set $C$ with a substitution probability noted as $P_e$.
Through this way, we make one training set, one validation set and nine types of test sets. These test sets have different error distributions by controlling the keys and corresponding values of the used confusion set.

In practice, we first construct a large and sufficient confusion set $S$. 
$S$ is composed of two parts, one of which is phonetically similar confusion set $S_p$  and the other is graphically similar confusion set $S_g$.
We sample some keys of $S$ together with all corresponding values, as the confusion set of unseen error $S_{unseen}^{k}$, to create a test set in which the target characters have never made a mistake in the training set.
In the remaining confusion set $S-S_{unseen}^{k}$, we randomly select some keys and then sample some of their values as the confusion set $S_{unseen}^{v}$.  The role of $S_{unseen}^{v}$ is to create a test set in which the target characters have not made such types of errors in the training set.
Then the remaining confusion set $S-S_{unseen}^{k}-S_{unseen}^{v}$ also noted as $S_{train}$ is used to build training set and validation set. 
Expressed in mathematical terms:
$$\forall k^{\prime} \in S_{unseen}^{k}, k^{\prime} \not\in S_{train}$$
$$\forall k^{\prime} \in S_{unseen}^{v}, S_{unseen}^{v}[k^{\prime}] \cup S_{train}[k^{\prime}]=\phi$$

\begin{figure*}
\centering 
\includegraphics[width=0.8\textwidth]{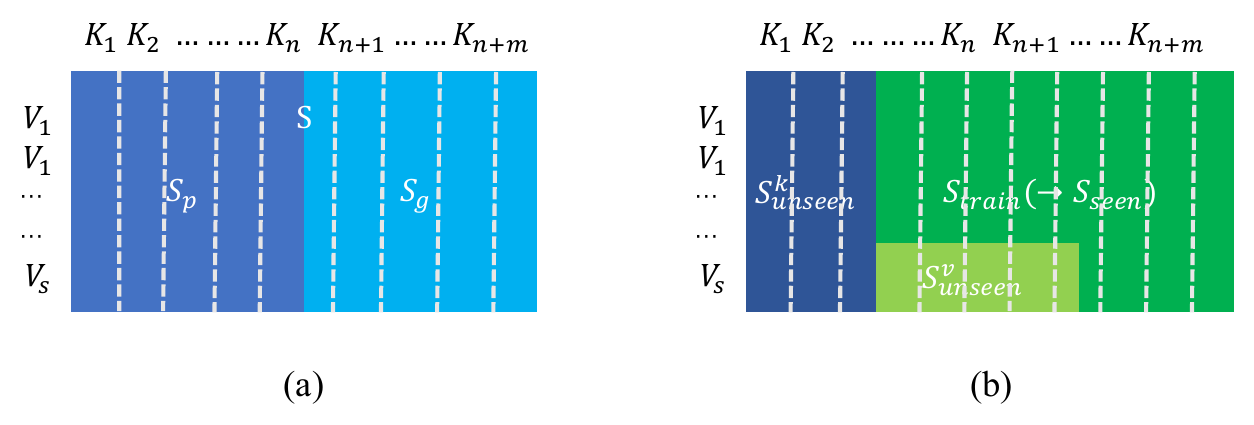} 
\caption{A visual representation of confusion set division.
(a) is the data structure and composition of the total confusion set $S$. 
(b) shows how to divide $S_{unseen}^k$, $S_{unseen}^v$ and $S_{train}$. And $S_{seen}$ is a subset of $S_{train}$.
It should be noted that all three confusion sets contain both phonetic and graphic confusion characters.} 
\label{splitvisual}
\end{figure*}

To measure the effect of the seen error, we select some of the misspelled character pairs that appear in the training set and note it as $S_{seen}$.
A visual representation of confusion set division is shown in Figure~\ref{splitvisual}.

As shown in Table \ref{tab:dataset}, we manufacture one training set, one validation set and nine types of test sets with different properties by controlling confusion sets. 
Among these test sets, the relatively special one is \emph{SContext}, whose purpose is to measure the effect of the seen context. It is made by sampling 5K sentences from \emph{Trainset} and replacing every wrong character $v_{ij}$ in source sentence with another error-prone candidate $v_{ik}$ of the same target character $k_i$ according to  confusion set $S$.
All datasets are made with 5\% substitution probability using the corresponding confusion set, except for \emph{Probs} and \emph{UnseenK}. 
\emph{Probs} is intended to evaluate the performance of the model in the face of sentences with different error frequencies.
So a series of test sets with different substitution probabilities are created.
It is worth pointing out that the number of keys of the used confusion set affects the frequency of error occurrence in the sentence with the same $P_e$, and the number of misspelling pairs affects the diversity of errors.
As for \emph{UnseenK}, since the key of $S_{unseen}^k$ is too few, it may lead to too low error frequency of the sentence. We use $P_e^{UnseenK}$ = 15\% so that $P_e^{UnseenK} \times N_{Sk} \approx P_e^{UnseenV} \times N_{Sv}$, where $N_{Sk}$ and $N_{Sv}$ mean the number of keys of $S_{unseen}^k$ and $S_{unseen}^v$ respectively.
Details of each confusion set are shown in Table~\ref{tab:confuset}.

\begin{table}[]
\small
\centering
\resizebox{0.47\textwidth}{!}{%
\begin{tabular}{c|c|cc}
\toprule
\textbf{Dataset}               & \textbf{Name}                              & \textbf{Confusion Sets}   & \textbf{Notes}                                            \\ \midrule
Training              & \emph{Trainset}  & $S_{train}$      & $S_{train} \subset S$                            \\ \midrule
Validation            & \emph{Validset}  & $S_{train}$      &                                                  \\ \midrule
\multirow{9}{*}{Test} & \emph{Regular}   & $S$              & $S=S_p \cup S_g$                                 \\  
                      & \emph{Probs}     & $S$              & Test sets with various $P_e$                     \\  
                      & \emph{Phonetics} & $S_p$            &                                                  \\  
                      & \emph{Graphics}  & $S_g$            &                                                  \\  
                      & \emph{SError}    & $S_{seen}$       & $S_{seen} \subset S_{train}$                     \\  
                      & \emph{SContext}  & $S$              & Same context as \emph{Trainset} \\  
                      & \emph{UnseenK}   & $S_{unseen}^{k}$ &                                                  \\  
                      & \emph{UnseenV}   & $S_{unseen}^{v}$ &                                                  \\ 
                      & \emph{Correct}   & $\phi$           & All are correct sentences                        \\ \bottomrule
\end{tabular}%
}
\caption{Datasets and the way they are made.}
\label{tab:dataset}
\end{table}

% \subsection{Details of Datasets Construction}
\begin{table}[]
\centering
\resizebox{0.5\textwidth}{!}{%
\begin{tabular}{c c c}
\toprule
Confusion   Set  & Number of Keys & Number of Error Pairs \\ \midrule
$S$              & 5303           & 216483                \\ 
$S_p$            & 5285           & 143972                \\ 
$S_g$            & 5296           & 80872                 \\ 
$S_{train}$      & 4075           & 166779                \\ 
$S_{unseen}^{k}$ & 1228           & 26149                 \\ 
$S_{unseen}^{v}$ & 3990           & 24498                 \\ \bottomrule
\end{tabular}%
}
\caption{Details of the size of each confusion set. Confusion set  $S_p$ and $S_g$ have duplicate error pairs so $S$ is not just adding up $S_p$ and $S_g$ and the error pairs number of $S$ is less than the sum of $S_{unseen}^{k}$, $S_{unseen}^{v}$ and $S_{train}$.}
\label{tab:confuset}
\end{table}

\begin{table*}[]
\small
\centering
\resizebox{0.98\textwidth}{!}{%
\begin{tabular}{c|c|cccc|cccc}
\toprule
\multirow{2}{*}{\textbf{Information}} & \multirow{2}{*}{\textbf{Model}} & \multicolumn{4}{c|}{\textbf{Detection Level}(\%)} & \multicolumn{4}{c}{\textbf{Correction Level}(\%)} \\ \  
                             &                        & Acc.      & Pre.      & Rec.      & F1      & Acc.      & Pre.      & Rec.      & F1      \\ \midrule
None                         & BERT                   & 82.26    & 80.8     & 78.94    & 79.86   & 65.54    & 59.8     & 58.42    & 59.1    \\ \midrule
\multirow{3}{*}{Phonetic}    & BT-PSym-AE             & 68.28    & 65.61    & 62.1     & 63.81   & 52.66    & 45.36    & 42.93    & 44.11   \\   
                             & BT-PSym                & \textbf{83.52}    & \textbf{81.94}    & \textbf{80.54}    & \textbf{81.23}   & 69.14    & 63.99    & 62.89    & 63.43   \\   
                             & BT-PMod                & 83.38    & 81.9     & 80.19    & 81.04   & 68.4     & 63.12    & 61.81    & 62.46   \\ \midrule
\multirow{3}{*}{Graphic}     & BT-GSym-AE             & 68.48    & 65.87    & 62.54    & 64.17   & 52.78    & 45.58    & 43.27    & 44.4    \\   
                             & BT-GSym               & 80.92    & 79.44    & 77.49    & 78.45   & 64.3     & 58.53    & 57.09    & 57.8    \\   
                             & BT-GMod                & 81.62    & 79.82    & 78.62    & 79.21   & 67.96    & 62.8     & 61.86    & 62.32   \\ \midrule
\multirow{2}{*}{Both}        & BT-PG                  & 83.08    & 81.63    & 79.97    & 80.79   & \textbf{69.7}     & \textbf{64.87}    & \textbf{63.55}    & \textbf{64.2}    \\   
                             & BT-PG-EG                & 82.4     & 80.78    & 79.04    & 79.9    & 68.68    & 63.57    & 62.2     & 62.88   \\ \bottomrule
\end{tabular}%
}
\caption{Performances of all models on \emph{Regular}, where accuracy (Acc.), precision (Prec.), recall (Rec.), F1 on sentence level are reported (\%). Best results are in \textbf{bold}.}
\label{tab:regularr}
\end{table*}

\section{Results and Analysis}

\subsection{Evaluation Methods}
We analyze the results mainly with the sentence-level metrics.
The results are reported at both detection level and correction level. 
At the detection level, a sentence is considered correct if all spelling errors in the sentence are successfully detected. 
At the correction level, the spelling errors not only need to be detected, but also need to be corrected. 
We report accuracy, precision, recall, and F1 scores for both levels.
To facilitate the analysis, we also calculate the detection and correction levels' perfomance scores with the character-level metrics.
We train all of our models on the \emph{Trainset}, validate them on the \emph{validset} and test them on all the test sets. 
All scores are shown in Appendix \ref{sec:result}.

\subsection{Overall Performance of Models}
The test set \emph{Regular} we constructed is similar to the error distribution in the real world, because it contains phonetic and graphic similarity errors, and there are both errors seen and not seen by the model. 
Therefore, the results on it, as shown in Table~\ref{tab:regularr}, can reflect the overall performance of different models. 
We can see that BERT provides a strong baseline.
Using the right way to fuse information has a good effect on improving the model performance.

\paragraph{Information Type} 
It can be seen that when incorporating a single type of information, the highest score (F1 score 63.43 in correction level) is obtained by BT-PSym. 
And the general performance is also better for the models that fuse phonetic information.
Therefore, we can find that the phonetic information is more useful compared to the graphic information.

There are several possible reasons: 
1) Phonetic information can be easily represented by pinyin sequences, which can be naturally encoded by the widely used sequence-to-sequence model. In contrast, the representation of stroke sequences can easily lose the spatial location information.
2) There are recursive problems with the strokes. For example, 鹅(goose) is made up of 我(I) and 鸟(bird) while 我 is made up of 手(hand) and 戈(weapon).
3）Since the confusion set $S_p$ is larger than $S_g$, the phonetic errors in the test set are more diverse. Therefore, adding phonetic information will have a more significant improvement on the results.

\paragraph{Source of Information} 
We can also find that different sources of the same type of information have a significant impact on the results.
The information from the pre-trained models performs well, especially the VGG with font images as input, which makes good use of graphic information.
It should be due to the knowledge learned by the model during pre-training.
BT-PSym using Transformer to encode pinyin sequences also performs surprisingly well, which is consistent with the conclusion of recent works \citep{zhang2021correcting,wang2021dynamic}.
BT-GSym performs average, probably because the complexity of the strokes makes it difficult to use them effectively.

\paragraph{Information Fusion Method}
The information fusion method has a high influence on the results. 
The difference between BT-PSym-AE and BT-PSym is that the former first sums the embeddings of the different inputs and then encodes them, while BT-PSym does in the opposite order.
The results of BT-PSym-AE are even lower than the original BERT, probably because adding up vector embeddings in different spaces makes the model confusing. 
And BT-PG is better than BT-PG-EG, indicating that the gating mechanism is not as good as concatenating the encoded information vectors and then feeding it to the transform layer.
This may be due to the fact that the use of transform layers allows more direct manipulation of the information compared to the gating mechanism that derives three weight values to sum the information vectors.

\subsection{Effect of Error Frequency}
\label{probsec}
\begin{figure}
\centering 
\includegraphics[width=0.48\textwidth]{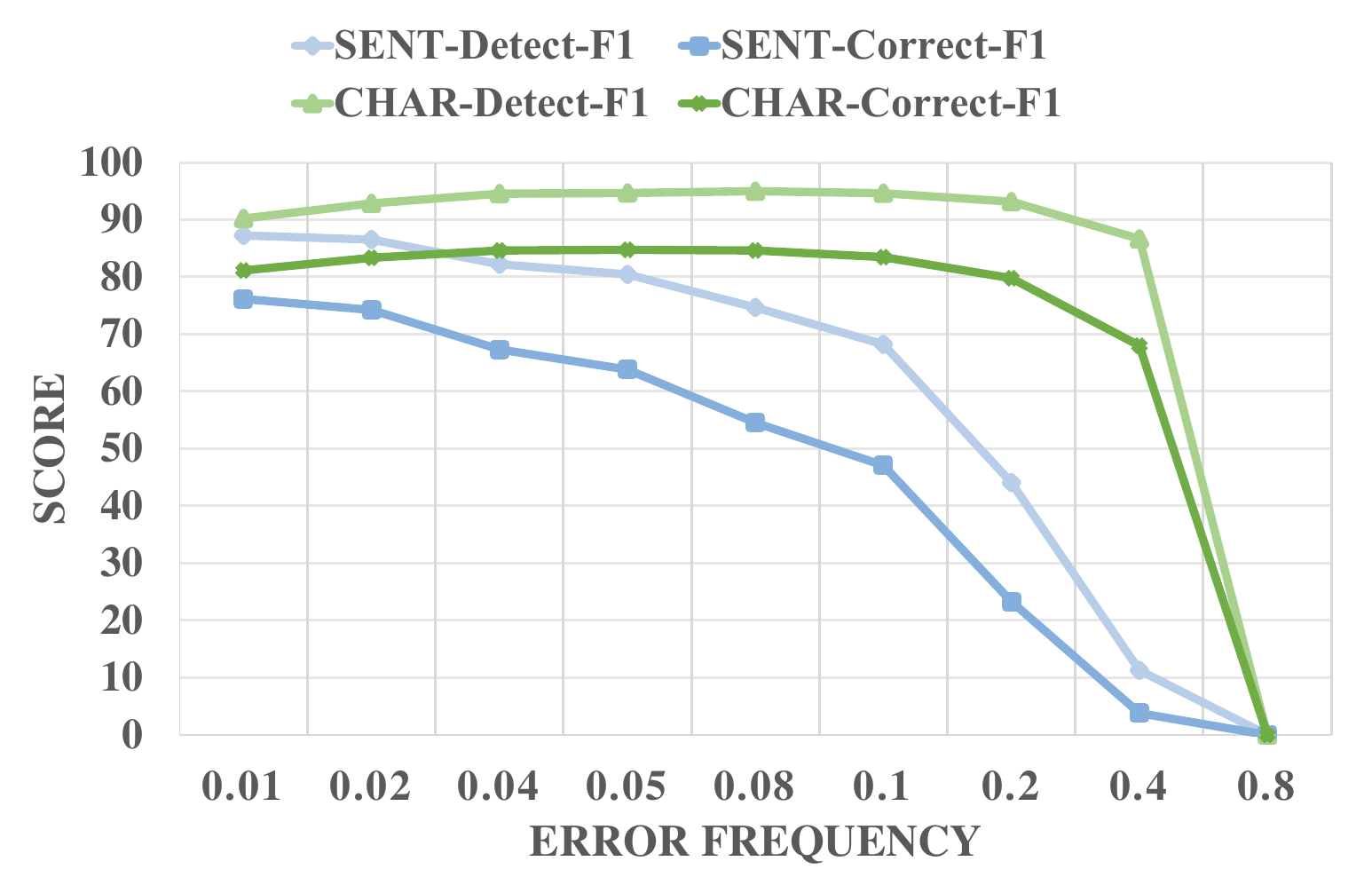} 
\caption{The performance of BT-PG when faced with different error frequencies in the sentences. CAHR-Detect-F1 and CHAR-Correct-F1 get the highest point at frequency of 5\%, and the highest point for SENT-Detect-F1 and SENT-Correct-F1 occurs at 1\%.} 
\label{Fig.3}
\end{figure}

Due to the limitation of space and without loss of generality, we choose BT-PG as a representative to analyze the performance of the model when facing different error frequencies in the sentences, as shown in Figure~\ref{Fig.3}. 
%The results of other models are similar that are shown in Appendix~\ref{sec:result}.

We can find that the sentence-level metrics are more affected by the frequency of errors, and the scores drop significantly with increasing frequency.
It is reasonable because the sentence-level metrics consider that all errors in a sentence need to be detected or corrected.
So the more errors in a sentence, the more difficult it is.

However, the character-level metrics perform steadily over a fairly large range of error frequencies (1\%-20\%) while frequency 20\% is bad enough in daily life.
It indicates that the model can still correct errors properly when faced with a range of different error frequencies, and that the sentence-level score decreases just because the model cannot correct the entire sentence completely. 
It is fun to note that the character-level metric is highest at 5\%, which is the frequency of the training set construction. It is possibly because that models are more adjusted to this frequency.

At lower error frequencies, the scores for the two levels are similar, which is not surprising since there may be only one error per sentence, so the sentence-level metrics degrade to the character-level ones.
The scores for the two are also similar for a high number of errors, because at that time the sentences are difficult to understand and models are confused and cannot perform error correction.

Therefore, the error frequency has a great impact on the performance of the model. The character-level metrics are more stable and are a more realistic reflection of the model's error correction capability, while almost all recent CSC work reports sentence-level results.

\subsection{Ability to Keep Correct}

\begin{table}[]
\centering
\resizebox{0.48\textwidth}{!}{%
\begin{tabular}{cc|cc}
\toprule
\textbf{Model}      & \textbf{Accuracy} & \textbf{Model}      & \textbf{Accuracy}                        \\ \midrule
BERT       & 0.981    &            &                                 \\ \midrule
BT-PSym-AE & 0.961    & BT-GSym-AE & 0.955                           \\ \midrule
BT-PSym    & 0.982    & BT-GSym    & 0.978                           \\ \midrule
BT-PMod    & 0.984    & BT-GMod    & 0.955                           \\ \midrule
BT-PG      & 0.981    & BT-PG-EG   & \textbf{0.985} \\ \bottomrule
\end{tabular}%
}
\caption{Accuracy on \emph{Correct}.}
\label{tab:correct}
\end{table}

To evaluate the ability of the models to not add errors in the face of correct sentence input, we construct \emph{Corrects} set whose source sentences are all correct sentences, and the results are shown in Table~\ref{tab:correct}.
We can see that all models maintain the correct sentences very well and do not add many new errors to them.
Meanwhile, this accuracy can be compared with the detection-level scores in \emph{Regular} set. The latter is significantly lower than the former, indicating that compared to correct sentences, sentences containing errors can mislead the model to incorrectly change other words in the sentence or ignore errors.

\subsection{Error Type and Information Type}
\label{sectype}
\begin{table}[]
\centering
\resizebox{0.48\textwidth}{!}{%
\begin{tabular}{ccccc}
\toprule
\multicolumn{1}{c}{\multirow{2}{*}{\textbf{Information}}} & \multicolumn{2}{|c|}{\textbf{SENT-Level}}                                  & \multicolumn{2}{c}{\textbf{CHAR-Level}}              \\ 
\multicolumn{1}{c}{}                             & \multicolumn{1}{|c}{detect-F1} & \multicolumn{1}{c|}{correct-F1} & \multicolumn{1}{c}{detect-F1} & correct-F1 \\  \midrule
\multicolumn{5}{c}{Test set: \emph{Phonetics}}                                                                                                                                                    \\ \midrule
\multicolumn{1}{c|}{None}                         & 83.62                          & 67.26                           & 95.66                          & 86.44                           \\ 
\multicolumn{1}{c|}{+Phonetic}                     & +1.96                           & +6.75                            & +0.38                           & +3.31                            \\ 
\multicolumn{1}{c|}{+Both}                         & +0.8                            & +6.25                            & +0.23                           & +2.95                            \\ \midrule
\multicolumn{5}{c}{Test set: \emph{Graphics}}                                                                                                                                                     \\ \midrule
\multicolumn{1}{c|}{None}                         & 76.24                          & 51.83                           & 93.06                          & 76.98                           \\ 
\multicolumn{1}{c|}{+Graphic}                      & +0.72                           & +5.11                            & +0.11                           & +3.29                            \\ 
\multicolumn{1}{c|}{+Both}                         & +0.33                           & +3.57                            & +0.21                           & +2.09                            \\ \bottomrule
\end{tabular}%
}
\caption{Results on \emph{Phonetics} and \emph{Graphics}}
\label{tab:pinyinshape}
\end{table}

In order to analyze the role of fusion information more clearly, we construct test sets based only on either type of error (i.e., \emph{Phonetics} or \emph{Graphics}) and the results are shown in Table~\ref{tab:pinyinshape}.
We can see that for a certain kind of error, fusing the corresponding information has a facilitating effect, while further fusing another kind of information has a slightly negative effect.
It also proves that our models do learn to use the corresponding information effectively.

The fused information does not improve the detection score much, but it improves the correction score significantly. 
This may be because as a masked language model, the model can easily diagnose where there are errors, but without any hints of phonetic or graphic information, it would only predict the words with the highest probability.

\begin{figure}[h]
\centering 
\includegraphics[width=0.48\textwidth]{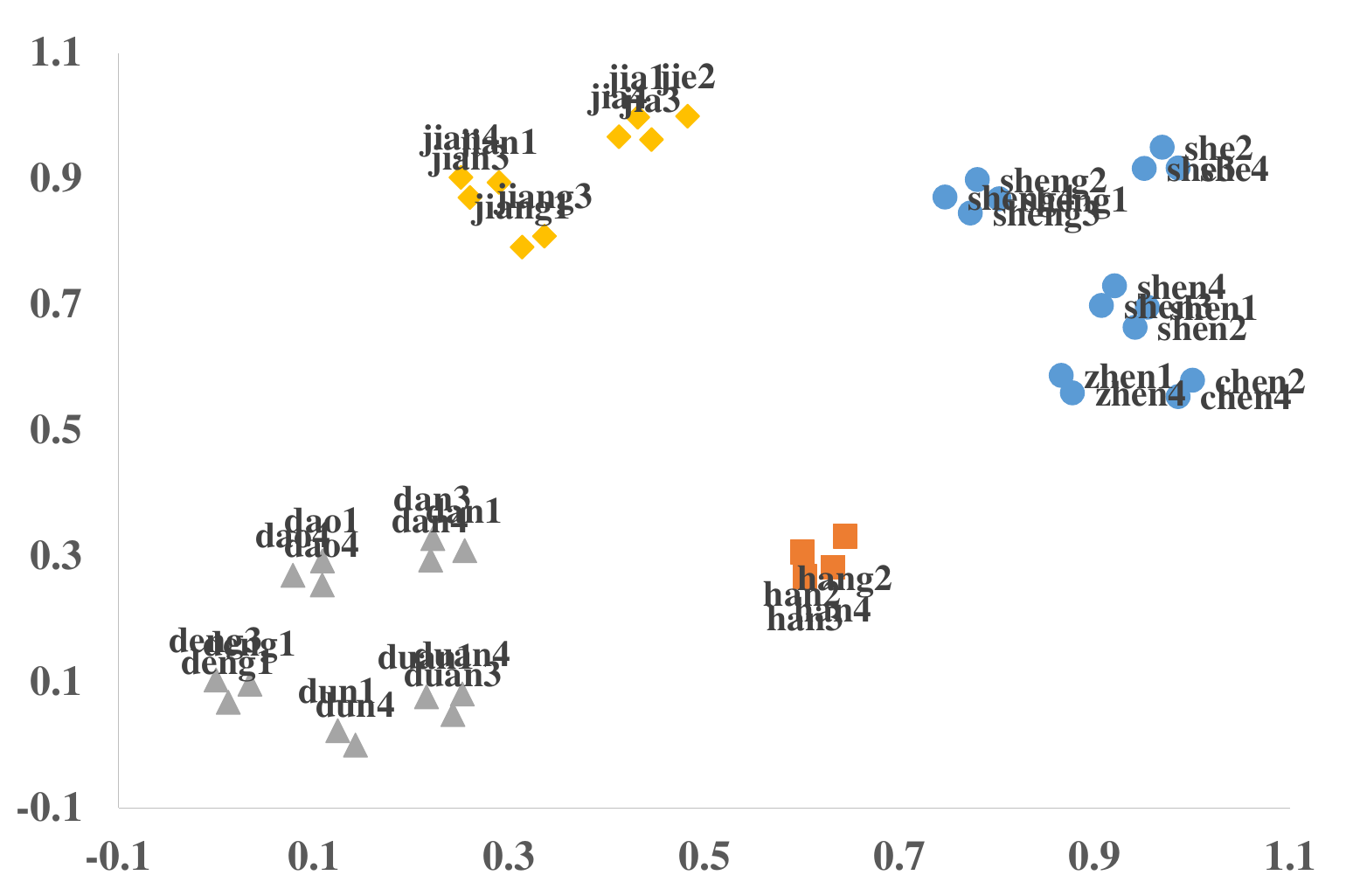} 
\caption{The scatter of phonetic features after dimensionality reduction. Characters with the similar pronunciation, which are reflected through pinyin, are clustered together.} 
\label{Fig pinyin}
\end{figure}

\begin{figure}[h]
\centering 
\includegraphics[width=0.48\textwidth]{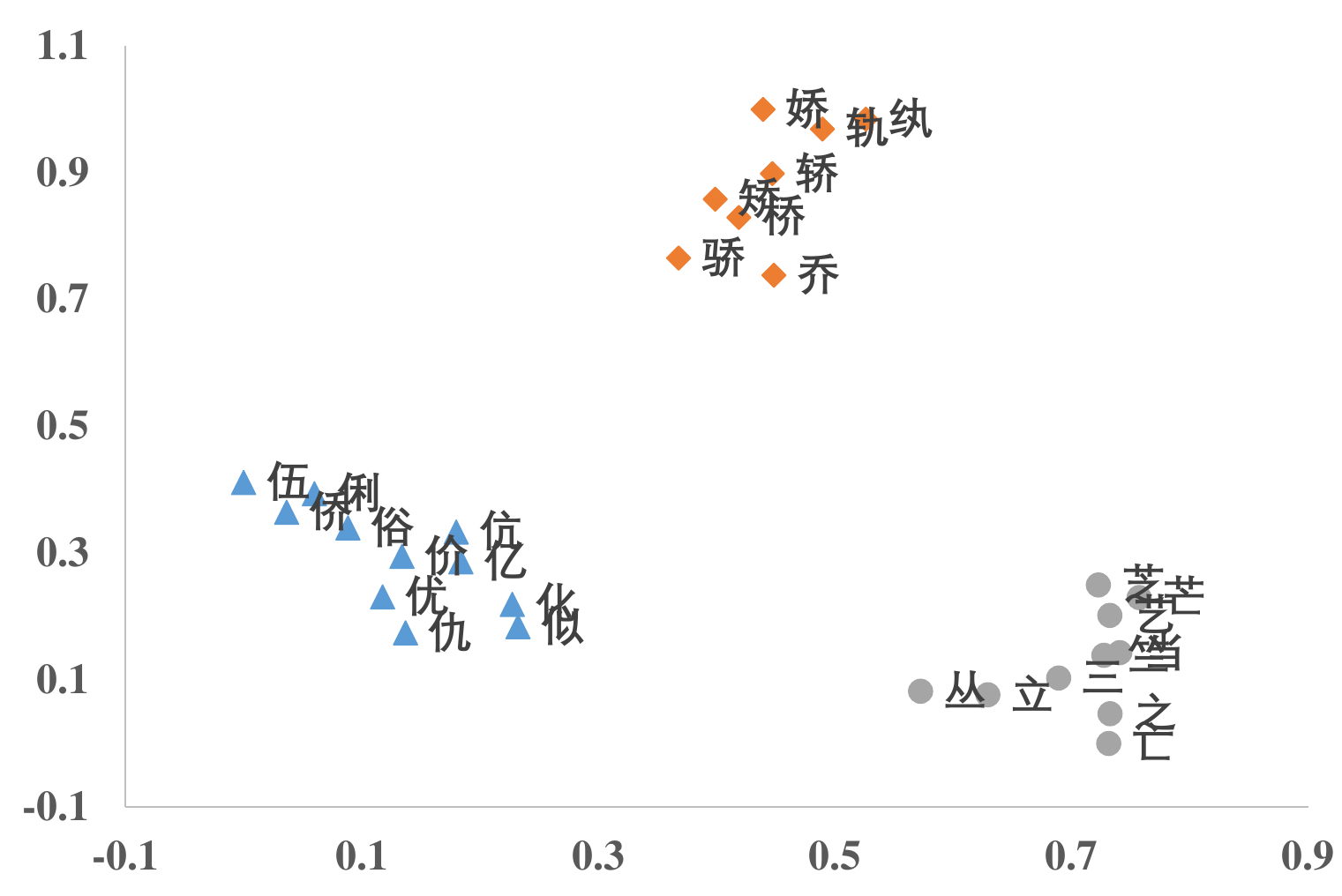} 
\caption{The scatter of graphic features after dimensionality reduction. Characters that are clustered together have similar strokes or structures.} 
\label{Fig shape}
\end{figure}

To understand the effectiveness of the fused information more intuitively, we choose some character examples, perform the dimensionality reduction on the phonetic and graphic features from BT-PG and visualize them using t-SNE \citep{van2008visualizing}. 
The input to BT-PG is the font images of the characters and the synthesized speech features. The results are shown in Figure~\ref{Fig pinyin} and Figure~\ref{Fig shape}. 
It further validates the effectiveness of phonetic and graphic information.

\subsection{Unseen Errors}

\begin{figure}
\centering 
\includegraphics[width=0.48\textwidth]{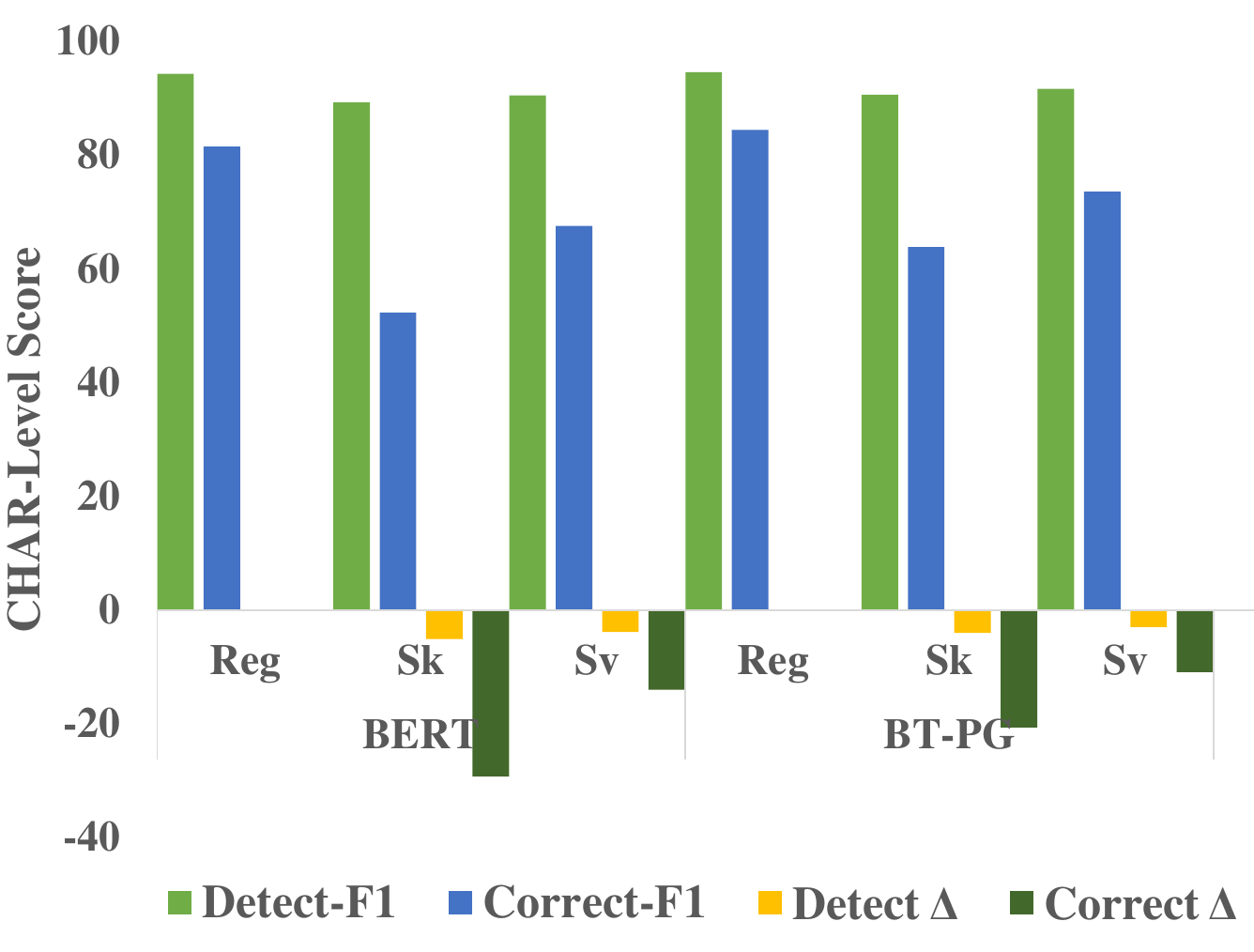} 
\caption{Character-level scores for BERT and BT-PG on \emph{Regular}(Reg), \emph{UnseenK}(Sk), and \emph{UnseenV}(Sv). Detect $\Delta$ and Correct $\Delta$ mean the score of the test set minus the score of \emph{Regular}.} 
\label{Fig unseen}
\end{figure}

The results on \emph{UnseenK} and \emph{UnseenV} are shown in Figure~\ref{Fig unseen}.
We can find that the scores of all the models drop substantially in the face of unseen errors. 
This also tells us that the existing models are not yet good at correcting unknown errors, and there is still much room for improvement in the CSC task. 
However, we can also see that BT-PG has a smaller drop compared to BERT, which indicates that the phonetic information and the graphic information do help the model understand the CSC task and improve the generalization ability of the model in the face of unknown errors.

Also, we can find higher scores in \emph{UnseenV} than in \emph{UnseenK}. 
For the orthographic-misspelling pair $(a,b)$ in \emph{UnseenV}, the model has never seen that $b$ can be changed to $a$ during training. 
While for the $(a,b)$ pair in \emph{UnseenK}, the model has never seen that any word can be changed to $a$.
It inspires us to make the confusion set contain more key values when data augmentation is applied to give the model a richer experience for errors.

As in the previous Section~\ref{sectype}, models perform consistently in detection level, which should be benefited from the way BERT is pre-trained.

\subsection{Seen Errors and Seen Contexts}
\label{seensec}
\begin{table}[]
\centering
\resizebox{0.4\textwidth}{!}{%
\begin{tabular}{c|cc|cc} 
\toprule
\multirow{2}{*}{\textbf{Model}} & \multicolumn{2}{c|}{\textbf{Seen Context}} & \multicolumn{2}{c}{\textbf{Seen Error}}   \\
                                & Dec-F1 & Cor-F1   & Dec-F1 & Cor-F1  \\ 
\midrule
BERT                            & 97.27              & 94.15                 & 96.4               & 92.58                \\
BT-PG                           & 97.15              & 94.65                 & 96.07              & 92.7                 \\
\bottomrule
\end{tabular}
}
\caption{Character-level results on \emph{SContext} and \emph{SError}.}
\label{tab:seen}
\end{table}

How much do seen contexts and seen errors affect the performance of the model? The results are shown in Table~\ref{tab:seen}.
We can see that models' scores very high and corrects almost all errors.
This means that the model can easily correct errors if it has seen a context similar to the test sentence. 
Similarly, if the model has learned this error situation, it will easily correct it too.

It inspires us, on the one hand, to let confusion set cover as many errors as possible when performing data augmentation.
On the other hand, we should construct training data using texts from the same domain according to task situations, with the aim that the model can see more similar contexts. 
The former has been used in recent work, while the latter points to a promising direction to explore.

\section{Analysis of SIGHAN Datasets}

% Please add the following required packages to your document preamble:
% \usepackage{graphicx}
\begin{table*}[]
\centering
\resizebox{0.79\textwidth}{!}{%
\begin{tabular}{c|ccccc}
\toprule
\textbf{Dataset}     & \textbf{\#Sent}                     & \textbf{\#Error} & \textbf{\#Error-pair} & \textbf{SIGHANTrain}\% & \textbf{Wang271K}\% \\ \midrule
SIGHAN13    & 1000                       & 1217    & 748          & 32.5\%        & 96.1\%     \\ \midrule
SIGHAN14    & 1062                       & 769     & 461          & 60.1\%        & 95.9\%     \\ \midrule
SIGHAN15    & 1100                       & 703     & 460          & 56.3\%        & 96.5\%     \\ \midrule
SIGHANTrain & 6126                       & 8470  & 3318         &             &          \\ \midrule
Wang271K    & \multicolumn{1}{l}{271329} & 381962    & 22409        &             &          \\ \bottomrule
\end{tabular}%
}
\caption{Statistics of the SIGHAN (transferred to simplified Chinese) and Wang271K. Columns SIGHANTrain\% and Wang271K\% mean the ratio of Error pairs in the test set that are covered by SIGHANTrain and Wang271K.}
\label{tab:SIGHAN}
\end{table*}

As shown in Table~\ref{tab:SIGHAN}, we also conduct an analysis of SIGHAN datasets and the generated pseudo data by \citet{wang2019confusionset} (denoted as Wang271K), which are the most commonly used datasets in CSC task.
SIGHAN datasets have some critical drawbacks:
1) The whole dataset is too small, with only a few thousand sentence pairs in the training set and limited errors in the test sets.
2) For the reasons above, confusion sets used for data augmentation can easily cover the errors in SIGHAN test sets. So the results on SIGHAN can not credibly reflect the real error correction ability of the model.
3) SIGHAN datasets are in traditional Chinese, and most of the contemporary research is in simplified Chinese. Although there are some tools such as OpenCC\footnote{\url{https://github.com/BYVoid/OpenCC}} to convert traditional Chinese to simplified Chinese, some parts of the data are still not compatible with the simplified Chinese habit.
4) There is some noise in SIGHAN datasets, for example some errors are not corrected. Examples and details can be found in Appendix \ref{sec:sighan}.

In the meanwhile, it is worth pointing out that the data augmentation set Wang271K covers almost all the error pairs that appear in SIGHAN test sets. 
According to the discussion in Section~\ref{seensec}, it can significantly improve the score on SIGHAN. 
So we think such evaluation method is not fair.
To prove it more convincingly, we train BART \citep{lewis2019bart} on the SIGHAN training set and Wang271K respectively, and test it on the SIGHAN test sets. We find a significant improvement of about 15 points in the results. However, there is no significant improvement on the other test set we constructed. The details and results are shown in Appendix \ref{sec: bart}.
Considering that almost all work is currently using Wang271K as extra dataset, we believe that SIGHAN can not fairly and credibly reflect the performance of the model.
The high score on SIGHAN now does not mean that the CSC task has made satisfactory progress.

\section{Conclusion}
In this paper we conducted a comprehensive analysis study for CSC by building a variety of test sets and implementing typical CSC models. Our evaluation concludes that the introduction of phonetic and graphic information has a significant effect on CSC, but the current model still performs poorly against unseen errors. 
The error distribution of the test set also has a significant impact on the performance of the model. 
Evaluations on the commonly used SIGHAN datasets are not credible and there is still much room for exploration and progress in the CSC task.

\section*{Limitations}
We only show the results of training the model on the training set with a substitution probability of 5\%. 
Although we also conduct experiments on training sets with other substitution probabilities and obtain the same conclusions as in the paper, we still do not fully explore the impact of the training set due to the space limitation of the paper and the large number of models and test sets we constructed.
For example, training sets containing only phonetic similarity errors or graphic similarity errors are not constructed.
These experiments can be explored in future work.

\section*{Ethics Statement}
The corpus we use is open source official Chinese news articles, which do not include any racist, sexist, hate speech or other toxic language. The Chinese characters in our confusion set are also commonly used characters.

% \section*{Acknowledgements}
% This document has been adapted by Yue Zhang, Ryan Cotterell and Lea Frermann from the style files used for earlier ACL and NAACL proceedings, including those for 
% ACL 2020 by Steven Bethard, Ryan Cotterell and Rui Yan,
% ACL 2019 by Douwe Kiela and Ivan Vuli\'{c},
% NAACL 2019 by Stephanie Lukin and Alla Roskovskaya, 
% ACL 2018 by Shay Cohen, Kevin Gimpel, and Wei Lu, 
% NAACL 2018 by Margaret Mitchell and Stephanie Lukin,
% Bib\TeX{} suggestions for (NA)ACL 2017/2018 from Jason Eisner,
% ACL 2017 by Dan Gildea and Min-Yen Kan, NAACL 2017 by Margaret Mitchell, 
% ACL 2012 by Maggie Li and Michael White, 
% ACL 2010 by Jing-Shin Chang and Philipp Koehn, 
% ACL 2008 by Johanna D. Moore, Simone Teufel, James Allan, and Sadaoki Furui, 
% ACL 2005 by Hwee Tou Ng and Kemal Oflazer, 
% ACL 2002 by Eugene Charniak and Dekang Lin, 
% and earlier ACL and EACL formats written by several people, including
% John Chen, Henry S. Thompson and Donald Walker.
% Additional elements were taken from the formatting instructions of the \emph{International Joint Conference on Artificial Intelligence} and the \emph{Conference on Computer Vision and Pattern Recognition}.

% % Entries for the entire Anthology, followed by custom entries
% \bibliography{anthology,custom}
% \bibliographystyle{acl_natbib}

% Entries for the entire Anthology, followed by custom entries
\bibliography{anthology,custom}
\bibliographystyle{acl_natbib}

\clearpage

\appendix
\section{Aspects of  Different Models Compared}
\begin{figure*}
\centering 
\includegraphics[width=0.95\textwidth]{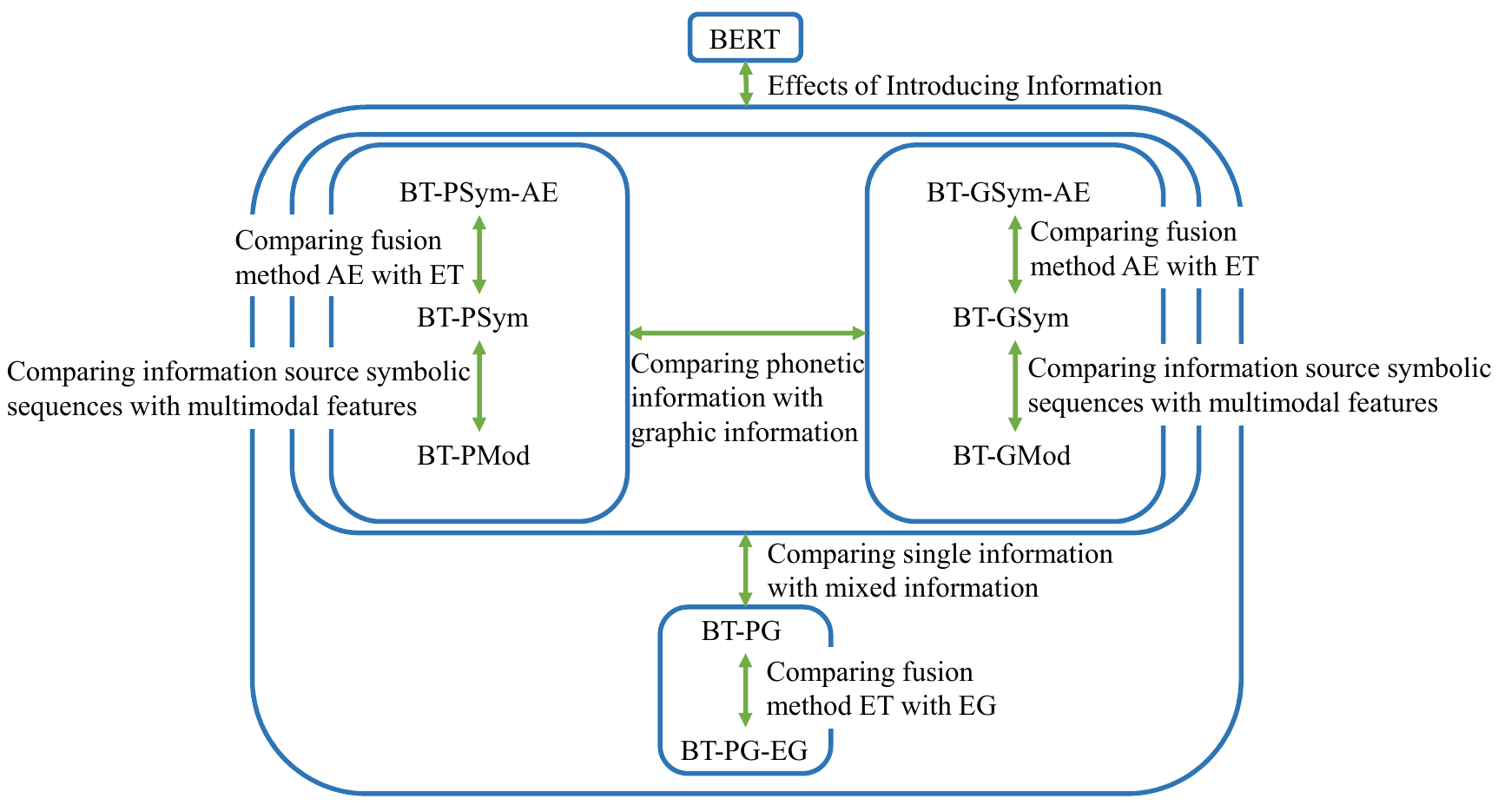} 
\caption{Comparison aspects of different models. The comparison aspects can be divided into three main types: the source of information introduced, the type of information introduced, and the way of fusing information.} 
\label{modelcomp}
\end{figure*}

The aspects of the different models compared are shown in Figure~\ref{modelcomp}. 
\label{modelcompa}

\section{Implementation Details of Models}
\label{detailmodel}
We implement our models using PyTorch framework \citep{paszke2019pytorch} with the Transformers library \citep{wolf2020transformers}.
To ensure a fair and scientific comparison, the implementation of the same functional part is the same for all models. The semantic information module is initialized by BERT \citep{devlin2018bert} and the generation module is a classifier implemented by MLP.
For the encoder in Figure~\ref{fuse}, we set the number of Transformer encoder layers to 4.
For the GMod, we collect one kind of the Chinese character fonts, namely Gothic typefaces. And we put these images into VGG19 and obtain its output vectors.
And we use the hidden representation from Tacotron2 and transform it into the dimension of 768.
All the embeddings and hidden states have the dimension of 768, same as BERT.
We train all models with the AdamW optimizer.
The learning rate is set to 5e-5 and all models are trained with learning rate warming up and linear decay.
Our other hyperparameters and evaluation codes are based on \citet{xu2021read}\footnote{\url{https://github.com/DaDaMrX/ReaLiSe}}.
The number of parameters of our models are similar to BERT-base.
We train all models on 8 Tesla K80 GPU for two days.

% \section{Appendix: A Visual Representation of Confusion Set Division}
% \label{splitvisuala}
% The visual representation of confusion set division is shown in Figure~\ref{splitvisual}.

\section{Appendix: All the Results of Models}
We experiment the nine models on nine test sets and calculate sentence-level and character-level scores. The results on \emph{Correct} and \emph{Probs} have been shown in the main article. The detailed results on the other test sets are shown in the following tables \ref{rc}-\ref{sec}.
\label{sec:result}

% Please add the following required packages to your document preamble:
% \usepackage{multirow}
% \usepackage{graphicx}
\begin{table*}[]
\centering
\resizebox{0.9\textwidth}{!}{%
\begin{tabular}{c|c|cccc|cccc}
\hline
\multirow{2}{*}{Information} & \multirow{2}{*}{Models} & \multicolumn{4}{c|}{Detection Level} & \multicolumn{4}{c}{Correction Level} \\ \cline{3-10} 
                             &                         & Acc.    & Pre.    & Rec.    & F1     & Acc.    & Pre.    & Rec.    & F1     \\ \hline
None                         & BERT                    & 92.34   & 96.92   & 91.94   & 94.36  & 80.91   & 83.78   & 79.47   & 81.57  \\ \hline
\multirow{3}{*}{Phonetic}    & BT-PSym-AE              & 83.46   & 94.95   & 82.37   & 88.22  & 69.67   & 77.63   & 67.35   & 72.13  \\ \cline{2-10} 
                             & BT-PSym                 & 93.06   & 96.88   & 92.74   & 94.77  & 83.31   & 85.78   & 82.11   & 83.91  \\ \cline{2-10} 
                             & BT-PMod                 & 92.81   & 97.03   & 92.39   & 94.65  & 82.8    & 85.57   & 81.48   & 83.48  \\ \hline
\multirow{3}{*}{Graphic}     & BT-GSym-AE              & 84.21   & 94.54   & 83.28   & 88.55  & 70.06   & 77.02   & 67.85   & 72.15  \\ \cline{2-10} 
                             & BT-GSym                 & 91.65   & 96.71   & 91.26   & 93.9   & 79.93   & 83.17   & 78.48   & 80.75  \\ \cline{2-10} 
                             & BT-GMod                 & 92.99   & 95.39   & 92.83   & 94.09  & 83.5    & 84.75   & 82.48   & 83.6   \\ \hline
\multirow{2}{*}{Both}        & BT-PG                   & 92.82   & 96.92   & 92.47   & 94.64  & 83.72   & 86.53   & 82.55   & 84.49  \\ \cline{2-10} 
                             & BT-PG-EG                & 92.21   & 97.14   & 91.76   & 94.37  & 82.82   & 86.31   & 81.53   & 83.85  \\ \hline
\end{tabular}%
}
\caption{Models' results on \emph{Regular} set on character-level metrics.}
\label{rc}
\end{table*}

% Please add the following required packages to your document preamble:
% \usepackage{multirow}
% \usepackage{graphicx}
\begin{table*}[]
\centering
\resizebox{0.9\textwidth}{!}{%
\begin{tabular}{c|c|cccc|cccc}
\hline
\multirow{2}{*}{Information} & \multirow{2}{*}{Models} & \multicolumn{4}{c|}{Detection Level} & \multicolumn{4}{c}{Correction Level} \\ \cline{3-10} 
                             &                         & Acc.    & Pre.    & Rec.    & F1     & Acc.    & Pre.    & Rec.    & F1     \\ \hline
None                         & BERT                    & 85.8    & 84.08   & 83.16   & 83.62  & 72.68   & 67.63   & 66.89   & 67.26  \\ \hline
\multirow{3}{*}{Phonetic}    & BT-PSym-AE              & 71.46   & 68.99   & 65.65   & 67.28  & 57.68   & 51.03   & 48.56   & 49.76  \\ \cline{2-10} 
                             & BT-PSym                 & 87.48   & 86.08   & 85.09   & 85.58  & 78.2    & 74.44   & 73.59   & 74.01  \\ \cline{2-10} 
                             & BT-PMod                 & 86.72   & 85.15   & 84.05   & 84.6   & 76.88   & 72.79   & 71.85   & 72.32  \\ \hline
\multirow{3}{*}{Graphic}     & BT-GSym-AE              & 72.38   & 69.67   & 66.84   & 68.23  & 58.44   & 51.65   & 49.55   & 50.58  \\ \cline{2-10} 
                             & BT-GSym                 & 84.84   & 83.02   & 81.94   & 82.48  & 71.38   & 66.11   & 65.25   & 65.68  \\ \cline{2-10} 
                             & BT-GMod                 & 84.38   & 82.17   & 81.85   & 82.01  & 73.72   & 68.9    & 68.63   & 68.76  \\ \hline
\multirow{2}{*}{Both}        & BT-PG                   & 86.88   & 85.4    & 84.42   & 84.91  & 77.74   & 73.93   & 73.09   & 73.51  \\ \cline{2-10} 
                             & BT-PG-EG                & 86      & 84.55   & 83.21   & 83.88  & 75.58   & 71.42   & 70.29   & 70.85  \\ \hline
\end{tabular}%
}
\caption{Models' results on \emph{Phonetics} set on sentence-level metrics.}
\label{ps}
\end{table*}

% Please add the following required packages to your document preamble:
% \usepackage{multirow}
% \usepackage{graphicx}
\begin{table*}[]
\centering
\resizebox{0.9\textwidth}{!}{%
\begin{tabular}{c|c|cccc|cccc}
\hline
\multirow{2}{*}{Information} & \multirow{2}{*}{Models} & \multicolumn{4}{c|}{Detection Level} & \multicolumn{4}{c}{Correction Level} \\ \cline{3-10} 
                             &                         & Acc.    & Pre.    & Rec.    & F1     & Acc.    & Pre.    & Rec.    & F1     \\ \hline
None                         & BERT                    & 94.49   & 97.09   & 94.27   & 95.66  & 86.19   & 87.74   & 85.18   & 86.44  \\ \hline
\multirow{3}{*}{Phonetic}    & BT-PSym-AE              & 86.26   & 95.33   & 85.37   & 90.08  & 74.84   & 81.37   & 72.86   & 76.88  \\ \cline{2-10} 
                             & BT-PSym                 & 95.27   & 97.04   & 95.06   & 96.04  & 89.59   & 90.69   & 88.84   & 89.75  \\ \cline{2-10} 
                             & BT-PMod                 & 94.96   & 97.13   & 94.69   & 95.89  & 88.71   & 90.1    & 87.84   & 88.95  \\ \hline
\multirow{3}{*}{Graphic}     & BT-GSym-AE              & 87.17   & 94.91   & 86.38   & 90.45  & 75.41   & 80.76   & 73.5    & 76.96  \\ \cline{2-10} 
                             & BT-GSym                 & 94.03   & 96.87   & 93.76   & 95.29  & 85.51   & 87.23   & 84.42   & 85.8   \\ \cline{2-10} 
                             & BT-GMod                 & 94.51   & 95.58   & 94.47   & 95.03  & 87.43   & 87.73   & 86.72   & 87.22  \\ \hline
\multirow{2}{*}{Both}        & BT-PG                   & 94.92   & 97.11   & 94.71   & 95.89  & 89.06   & 90.52   & 88.29   & 89.39  \\ \cline{2-10} 
                             & BT-PG-EG                & 94.49   & 97.14   & 94.19   & 95.64  & 88.03   & 89.85   & 87.12   & 88.46  \\ \hline
\end{tabular}%
}
\caption{Models' results on \emph{Phonetics} set on character-level metrics.}
\label{pc}
\end{table*}

% Please add the following required packages to your document preamble:
% \usepackage{multirow}
% \usepackage{graphicx}
\begin{table*}[]
\centering
\resizebox{0.9\textwidth}{!}{%
\begin{tabular}{c|c|cccc|cccc}
\hline
\multirow{2}{*}{Information} & \multirow{2}{*}{Models} & \multicolumn{4}{c|}{Detection Level} & \multicolumn{4}{c}{Correction Level} \\ \cline{3-10} 
                             &                         & Acc.    & Pre.    & Rec.    & F1     & Acc.    & Pre.    & Rec.    & F1     \\ \hline
None                         & BERT                    & 83.26   & 81.82   & 79.95   & 80.87  & 66.6    & 60.79   & 59.4    & 60.09  \\ \hline
\multirow{3}{*}{Phonetic}    & BT-PSym-AE              & 68.22   & 65.71   & 61.59   & 63.59  & 52.88   & 45.53   & 42.67   & 44.05  \\ \cline{2-10} 
                             & BT-PSym                 & 84.9    & 83.25   & 81.92   & 82.58  & 69.48   & 63.93   & 62.9    & 63.41  \\ \cline{2-10} 
                             & BT-PMod                 & 84.34   & 82.88   & 81.2    & 82.03  & 69.44   & 64.12   & 62.83   & 63.47  \\ \hline
\multirow{3}{*}{Graphic}     & BT-GSym-AE              & 69.4    & 66.77   & 63.2    & 64.93  & 53.6    & 46.18   & 43.71   & 44.91  \\ \cline{2-10} 
                             & BT-GSym                 & 82.58   & 80.92   & 79.18   & 80.04  & 65.74   & 59.69   & 58.41   & 59.05  \\ \cline{2-10} 
                             & BT-GMod                 & 82.44   & 80.26   & 79.13   & 79.69  & 68.94   & 63.37   & 62.48   & 62.92  \\ \hline
\multirow{2}{*}{Both}        & BT-PG                   & 84.06   & 82.35   & 80.91   & 81.62  & 70.64   & 65.5    & 64.36   & 64.92  \\ \cline{2-10} 
                             & BT-PG-EG                & 83.8    & 82.08   & 80.44   & 81.25  & 70.38   & 65.19   & 63.89   & 64.53  \\ \hline
\end{tabular}%
}
\caption{Models' results on \emph{Graphics} set on sentence-level metrics.}
\label{gs}
\end{table*}

% Please add the following required packages to your document preamble:
% \usepackage{multirow}
% \usepackage{graphicx}
\begin{table*}[]
\centering
\resizebox{0.9\textwidth}{!}{%
\begin{tabular}{c|c|cccc|cccc}
\hline
\multirow{2}{*}{Information} & \multirow{2}{*}{Models} & \multicolumn{4}{c|}{Detection Level} & \multicolumn{4}{c}{Correction Level} \\ \cline{3-10} 
                             &                         & Acc.    & Pre.    & Rec.    & F1     & Acc.    & Pre.    & Rec.    & F1     \\ \hline
None                         & BERT                    & 90.25   & 96.77   & 89.63   & 93.06  & 76.08   & 80.05   & 74.14   & 76.98  \\ \hline
\multirow{3}{*}{Phonetic}    & BT-PSym-AE              & 79.52   & 94.57   & 77.96   & 85.47  & 63.16   & 72.89   & 60.08   & 65.87  \\ \cline{2-10} 
                             & BT-PSym                 & 90.95   & 96.91   & 90.44   & 93.57  & 77.33   & 80.97   & 75.56   & 78.17  \\ \cline{2-10} 
                             & BT-PMod                 & 90.31   & 96.92   & 89.68   & 93.16  & 76.21   & 80.27   & 74.27   & 77.15  \\ \hline
\multirow{3}{*}{Graphic}     & BT-GSym-AE              & 80.84   & 93.95   & 79.51   & 86.13  & 63.92   & 72.09   & 61.01   & 66.09  \\ \cline{2-10} 
                             & BT-GSym                 & 89.45   & 96.6    & 88.81   & 92.54  & 74.81   & 79.2    & 72.81   & 75.87  \\ \cline{2-10} 
                             & BT-GMod                 & 91.43   & 95.32   & 91.11   & 93.17  & 79.89   & 82.12   & 78.49   & 80.27  \\ \hline
\multirow{2}{*}{Both}        & BT-PG                   & 90.63   & 96.74   & 90.05   & 93.27  & 78.09   & 82.01   & 76.34   & 79.07  \\ \cline{2-10} 
                             & BT-PG-EG                & 90.09   & 96.92   & 89.44   & 93.03  & 77.78   & 82.34   & 75.99   & 79.04  \\ \hline
\end{tabular}%
}
\caption{Models' results on \emph{Graphics} set on character-level metrics.}
\label{gc}
\end{table*}

% Please add the following required packages to your document preamble:
% \usepackage{multirow}
% \usepackage{graphicx}
\begin{table*}[]
\centering
\resizebox{0.9\textwidth}{!}{%
\begin{tabular}{c|c|cccc|cccc}
\hline
\multirow{2}{*}{Information} & \multirow{2}{*}{Models} & \multicolumn{4}{c|}{Detection Level} & \multicolumn{4}{c}{Correction Level} \\ \cline{3-10} 
                             &                         & Acc.    & Pre.    & Rec.    & F1     & Acc.    & Pre.    & Rec.    & F1     \\ \hline
None                         & BERT                    & 79.82   & 75.97   & 71.47   & 73.65  & 53.24   & 33.73   & 31.73   & 32.7   \\ \hline
\multirow{3}{*}{Phonetic}    & BT-PSym-AE              & 65.86   & 59.49   & 51.47   & 55.19  & 44.42   & 22.43   & 19.41   & 20.81  \\ \cline{2-10} 
                             & BT-PSym                 & 83.32   & 79.47   & 76.61   & 78.01  & 61.28   & 45.29   & 43.66   & 44.46  \\ \cline{2-10} 
                             & BT-PMod                 & 82.02   & 78.41   & 74.61   & 76.46  & 59.6    & 43.18   & 41.09   & 42.11  \\ \hline
\multirow{3}{*}{Graphic}     & BT-GSym-AE              & 66.64   & 60.42   & 52.99   & 56.46  & 44.94   & 23.42   & 20.54   & 21.89  \\ \cline{2-10} 
                             & BT-GSym                 & 78.58   & 73.97   & 69.68   & 71.76  & 52.04   & 31.84   & 29.99   & 30.89  \\ \cline{2-10} 
                             & BT-GMod                 & 80.28   & 75.5    & 72.7    & 74.07  & 58.06   & 40.99   & 39.47   & 40.22  \\ \hline
\multirow{2}{*}{Both}        & BT-PG                   & 81.68   & 77.66   & 74.1    & 75.84  & 61.04   & 45.31   & 43.24   & 44.25  \\ \cline{2-10} 
                             & BT-PG-EG                & 80.34   & 76.41   & 71.98   & 74.13  & 57.94   & 40.86   & 38.49   & 39.64  \\ \hline
\end{tabular}%
}
\caption{Models' results on \emph{UnseenK} set on sentence-level metrics.}
\label{us}
\end{table*}

% Please add the following required packages to your document preamble:
% \usepackage{multirow}
% \usepackage{graphicx}
\begin{table*}[]
\centering
\resizebox{0.9\textwidth}{!}{%
\begin{tabular}{c|c|cccc|cccc}
\hline
\multirow{2}{*}{Information} & \multirow{2}{*}{Models} & \multicolumn{4}{c|}{Detection Level} & \multicolumn{4}{c}{Correction Level} \\ \cline{3-10} 
                             &                         & Acc.    & Pre.    & Rec.    & F1     & Acc.    & Pre.    & Rec.    & F1     \\ \hline
None                         & BERT                    & 86.94   & 94.85   & 84.42   & 89.34  & 59.18   & 55.62   & 49.5    & 52.38  \\ \hline
\multirow{3}{*}{Phonetic}    & BT-PSym-AE              & 75.17   & 90.65   & 70.08   & 79.05  & 46.73   & 44.37   & 34.3    & 38.69  \\ \cline{2-10} 
                             & BT-PSym                 & 89.86   & 95.12   & 88.05   & 91.45  & 68.52   & 66.13   & 61.21   & 63.58  \\ \cline{2-10} 
                             & BT-PMod                 & 88.67   & 95.07   & 86.53   & 90.6   & 66.85   & 64.92   & 59.08   & 61.86  \\ \hline
\multirow{3}{*}{Graphic}     & BT-GSym-AE              & 76.28   & 90.25   & 71.65   & 79.88  & 48.19   & 45.75   & 36.32   & 40.5   \\ \cline{2-10} 
                             & BT-GSym                 & 86.14   & 94.4    & 83.46   & 88.59  & 57.6    & 53.79   & 47.55   & 50.48  \\ \cline{2-10} 
                             & BT-GMod                 & 88.89   & 92.71   & 87.17   & 89.85  & 65.7    & 61.68   & 57.99   & 59.78  \\ \hline
\multirow{2}{*}{Both}        & BT-PG                   & 89.02   & 94.74   & 86.96   & 90.68  & 68.61   & 66.77   & 61.29   & 63.92  \\ \cline{2-10} 
                             & BT-PG-EG                & 87.43   & 94.9    & 84.91   & 89.63  & 64.88   & 63.2    & 56.54   & 59.68  \\ \hline
\end{tabular}%
}
\caption{Models' results on \emph{UnseenK} set on character-level metrics.}
\label{uc}
\end{table*}

% Please add the following required packages to your document preamble:
% \usepackage{multirow}
% \usepackage{graphicx}
\begin{table*}[]
\centering
\resizebox{0.9\textwidth}{!}{%
\begin{tabular}{c|c|cccc|cccc}
\hline
\multirow{2}{*}{Information} & \multirow{2}{*}{Models} & \multicolumn{4}{c|}{Detection Level} & \multicolumn{4}{c}{Correction Level} \\ \cline{3-10} 
                             &                         & Acc.    & Pre.    & Rec.    & F1     & Acc.    & Pre.    & Rec.    & F1     \\ \hline
None                         & BERT                    & 73.26   & 70.37   & 67.5    & 68.91  & 51.34   & 41.89   & 40.18   & 41.02  \\ \hline
\multirow{3}{*}{Phonetic}    & BT-PSym-AE              & 53.92   & 49.41   & 43.82   & 46.45  & 32.92   & 19.9    & 17.65   & 18.71  \\ \cline{2-10} 
                             & BT-PSym                 & 76.7    & 74.11   & 71.78   & 72.93  & 55.32   & 46.6    & 45.14   & 45.86  \\ \cline{2-10} 
                             & BT-PMod                 & 74.46   & 71.67   & 68.97   & 70.29  & 53.42   & 44.42   & 42.75   & 43.57  \\ \hline
\multirow{3}{*}{Graphic}     & BT-GSym-AE              & 55.26   & 50.57   & 45.64   & 47.98  & 33.68   & 20.77   & 18.74   & 19.7   \\ \cline{2-10} 
                             & BT-GSym                 & 72.12   & 69.19   & 66.23   & 67.68  & 49.54   & 39.79   & 38.09   & 38.92  \\ \cline{2-10} 
                             & BT-GMod                 & 75.12   & 72.24   & 70.06   & 71.14  & 57.76   & 49.94   & 48.43   & 49.17  \\ \hline
\multirow{2}{*}{Both}        & BT-PG                   & 76.36   & 73.79   & 71.36   & 72.55  & 58.88   & 51.26   & 49.58   & 50.41  \\ \cline{2-10} 
                             & BT-PG-EG                & 75.78   & 73.35   & 70.51   & 71.9   & 58.04   & 50.35   & 48.4    & 49.36  \\ \hline
\end{tabular}%
}
\caption{Models' results on \emph{UnseenV} set on sentence-level metrics.}
\label{vs}
\end{table*}

% Please add the following required packages to your document preamble:
% \usepackage{multirow}
% \usepackage{graphicx}
\begin{table*}[]
\centering
\resizebox{0.9\textwidth}{!}{%
\begin{tabular}{c|c|cccc|cccc}
\hline
\multirow{2}{*}{Information} & \multirow{2}{*}{Models} & \multicolumn{4}{c|}{Detection Level} & \multicolumn{4}{c}{Correction Level} \\ \cline{3-10} 
                             &                         & Acc.    & Pre.    & Rec.    & F1     & Acc.    & Pre.    & Rec.    & F1     \\ \hline
None                         & BERT                    & 86.34   & 96.52   & 85.31   & 90.57  & 66.69   & 72.06   & 63.69   & 67.62  \\ \hline
\multirow{3}{*}{Phonetic}    & BT-PSym-AE              & 71.22   & 93.46   & 68.85   & 79.29  & 42.57   & 50.67   & 37.33   & 42.99  \\ \cline{2-10} 
                             & BT-PSym                 & 88.53   & 96.63   & 87.72   & 91.96  & 71.25   & 75.69   & 68.71   & 72.03  \\ \cline{2-10} 
                             & BT-PMod                 & 87.07   & 96.54   & 86.1    & 91.02  & 69.16   & 74.45   & 66.4    & 70.19  \\ \hline
\multirow{3}{*}{Graphic}     & BT-GSym-AE              & 72.88   & 92.87   & 70.73   & 80.3   & 44.17   & 51.41   & 39.15   & 44.45  \\ \cline{2-10} 
                             & BT-GSym                 & 85.51   & 96.05   & 84.46   & 89.88  & 65.04   & 70.44   & 61.94   & 65.92  \\ \cline{2-10} 
                             & BT-GMod                 & 88.55   & 94.81   & 87.84   & 91.19  & 74.43   & 78.05   & 72.31   & 75.07  \\ \hline
\multirow{2}{*}{Both}        & BT-PG                   & 88.38   & 96.36   & 87.55   & 91.74  & 74.38   & 79.41   & 72.15   & 75.6   \\ \cline{2-10} 
                             & BT-PG-EG                & 87.75   & 96.77   & 86.81   & 91.52  & 73.31   & 79.06   & 70.92   & 74.77  \\ \hline
\end{tabular}%
}
\caption{Models' results on \emph{UnseenV} set on character-level metrics.}
\label{vc}
\end{table*}

% Please add the following required packages to your document preamble:
% \usepackage{multirow}
% \usepackage{graphicx}
\begin{table*}[]
\centering
\resizebox{0.9\textwidth}{!}{%
\begin{tabular}{c|c|cccc|cccc}
\hline
\multirow{2}{*}{Information} & \multirow{2}{*}{Models} & \multicolumn{4}{c|}{Detection Level} & \multicolumn{4}{c}{Correction Level} \\ \cline{3-10} 
                             &                         & Acc.    & Pre.    & Rec.    & F1     & Acc.    & Pre.    & Rec.    & F1     \\ \hline
None                         & BERT                    & 91.42   & 90.57   & 89.68   & 90.12  & 86.67   & 84.57   & 83.74   & 84.15  \\ \hline
\multirow{3}{*}{Phonetic}    & BT-PSym-AE              & 77.82   & 75.18   & 73.2    & 74.18  & 69.68   & 64.74   & 63.03   & 63.87  \\ \cline{2-10} 
                             & BT-PSym                 & 91.2    & 90.14   & 89.34   & 89.74  & 87.23   & 85.14   & 84.38   & 84.76  \\ \cline{2-10} 
                             & BT-PMod                 & 90.74   & 89.78   & 88.79   & 89.28  & 86.76   & 84.76   & 83.83   & 84.29  \\ \hline
\multirow{3}{*}{Graphic}     & BT-GSym-AE              & 78.68   & 76.29   & 74.4    & 75.33  & 70.93   & 66.36   & 64.72   & 65.53  \\ \cline{2-10} 
                             & BT-GSym                 & 90.83   & 89.93   & 88.94   & 89.44  & 86.15   & 84.02   & 83.09   & 83.55  \\ \cline{2-10} 
                             & BT-GMod                 & 88.31   & 86.55   & 86.16   & 86.35  & 84.68   & 82      & 81.62   & 81.81  \\ \hline
\multirow{2}{*}{Both}        & BT-PG                   & 90.91   & 89.89   & 89.1    & 89.49  & 87.38   & 85.44   & 84.69   & 85.06  \\ \cline{2-10} 
                             & BT-PG-EG                & 90.49   & 89.36   & 88.45   & 88.9   & 86.89   & 84.81   & 83.95   & 84.38  \\ \hline
\end{tabular}%
}
\caption{Models' results on \emph{SContext} set on sentence-level metrics.}
\label{scs}
\end{table*}

% Please add the following required packages to your document preamble:
% \usepackage{multirow}
% \usepackage{graphicx}
\begin{table*}[]
\centering
\resizebox{0.9\textwidth}{!}{%
\begin{tabular}{c|c|cccc|cccc}
\hline
\multirow{2}{*}{Information} & \multirow{2}{*}{Models} & \multicolumn{4}{c|}{Detection Level} & \multicolumn{4}{c}{Correction Level} \\ \cline{3-10} 
                             &                         & Acc.    & Pre.    & Rec.    & F1     & Acc.    & Pre.    & Rec.    & F1     \\ \hline
None                         & BERT                    & 96.6    & 98.13   & 96.42   & 97.27  & 93.79   & 94.98   & 93.33   & 94.15  \\ \hline
\multirow{3}{*}{Phonetic}    & BT-PSym-AE              & 89.46   & 96.37   & 88.77   & 92.41  & 83.64   & 89.42   & 82.37   & 85.75  \\ \cline{2-10} 
                             & BT-PSym                 & 96.41   & 98.25   & 96.19   & 97.21  & 93.96   & 95.5    & 93.49   & 94.48  \\ \cline{2-10} 
                             & BT-PMod                 & 96.29   & 98.21   & 96.06   & 97.13  & 93.82   & 95.44   & 93.35   & 94.38  \\ \hline
\multirow{3}{*}{Graphic}     & BT-GSym-AE              & 90.19   & 96.11   & 89.62   & 92.75  & 84.49   & 89.38   & 83.34   & 86.25  \\ \cline{2-10} 
                             & BT-GSym                 & 96.36   & 98.15   & 96.15   & 97.14  & 93.54   & 94.99   & 93.05   & 94.01  \\ \cline{2-10} 
                             & BT-GMod                 & 96.05   & 96.73   & 95.96   & 96.34  & 93.64   & 94.06   & 93.31   & 93.68  \\ \hline
\multirow{2}{*}{Both}        & BT-PG                   & 96.38   & 98.12   & 96.2    & 97.15  & 94.13   & 95.59   & 93.72   & 94.65  \\ \cline{2-10} 
                             & BT-PG-EG                & 96.14   & 98.14   & 95.89   & 97     & 93.87   & 95.58   & 93.39   & 94.47  \\ \hline
\end{tabular}%
}
\caption{Models' results on \emph{SContext} set on character-level metrics.}
\label{scc}
\end{table*}

% Please add the following required packages to your document preamble:
% \usepackage{multirow}
% \usepackage{graphicx}
\begin{table*}[]
\centering
\resizebox{0.9\textwidth}{!}{%
\begin{tabular}{c|c|cccc|cccc}
\hline
\multirow{2}{*}{Information} & \multirow{2}{*}{Models} & \multicolumn{4}{c|}{Detection Level} & \multicolumn{4}{c}{Correction Level} \\ \cline{3-10} 
                             &                         & Acc.    & Pre.    & Rec.    & F1     & Acc.    & Pre.    & Rec.    & F1     \\ \hline
None                         & BERT                    & 89.1    & 87.59   & 87.08   & 87.34  & 83.72   & 80.82   & 80.35   & 80.58  \\ \hline
\multirow{3}{*}{Phonetic}    & BT-PSym-AE              & 76.34   & 73.8    & 71.36   & 72.56  & 69.2    & 64.56   & 62.43   & 63.48  \\ \cline{2-10} 
                             & BT-PSym                 & 88.88   & 87.5    & 86.73   & 87.12  & 83.94   & 81.26   & 80.55   & 80.91  \\ \cline{2-10} 
                             & BT-PMod                 & 88.6    & 87.22   & 86.31   & 86.76  & 83.66   & 80.98   & 80.13   & 80.55  \\ \hline
\multirow{3}{*}{Graphic}     & BT-GSym-AE              & 76.22   & 73.33   & 71.31   & 72.31  & 68.74   & 63.71   & 61.95   & 62.82  \\ \cline{2-10} 
                             & BT-GSym                 & 88.16   & 86.47   & 85.88   & 86.17  & 82.82   & 79.74   & 79.2    & 79.47  \\ \cline{2-10} 
                             & BT-GMod                 & 86.04   & 84.15   & 83.45   & 83.8   & 81.4    & 78.29   & 77.65   & 77.97  \\ \hline
\multirow{2}{*}{Both}        & BT-PG                   & 88.24   & 86.76   & 85.96   & 86.36  & 83.74   & 81.08   & 80.33   & 80.7   \\ \cline{2-10} 
                             & BT-PG-EG                & 88.2    & 86.66   & 85.68   & 86.17  & 83.82   & 81.11   & 80.2    & 80.65  \\ \hline
\end{tabular}%
}
\caption{Models' results on \emph{SError} set on sentence-level metrics.}
\label{ses}
\end{table*}

% Please add the following required packages to your document preamble:
% \usepackage{multirow}
% \usepackage{graphicx}
\begin{table*}[]
\centering
\resizebox{0.9\textwidth}{!}{%
\begin{tabular}{c|c|cccc|cccc}
\hline
\multirow{2}{*}{Information} & \multirow{2}{*}{Models} & \multicolumn{4}{c|}{Detection Level} & \multicolumn{4}{c}{Correction Level} \\ \cline{3-10} 
                             &                         & Acc.    & Pre.    & Rec.    & F1     & Acc.    & Pre.    & Rec.    & F1     \\ \hline
None                         & BERT                    & 95.9    & 97.04   & 95.78   & 96.4   & 92.46   & 93.19   & 91.98   & 92.58  \\ \hline
\multirow{3}{*}{Phonetic}    & BT-PSym-AE              & 88.91   & 95.18   & 88.17   & 91.54  & 83.02   & 88.18   & 81.69   & 84.81  \\ \cline{2-10} 
                             & BT-PSym                 & 95.72   & 97      & 95.54   & 96.27  & 92.41   & 93.3    & 91.9    & 92.59  \\ \cline{2-10} 
                             & BT-PMod                 & 95.51   & 97.04   & 95.29   & 96.16  & 92.25   & 93.38   & 91.7    & 92.53  \\ \hline
\multirow{3}{*}{Graphic}     & BT-GSym-AE              & 89.39   & 94.82   & 88.74   & 91.68  & 83.39   & 87.75   & 82.13   & 84.85  \\ \cline{2-10} 
                             & BT-GSym                 & 95.47   & 96.71   & 95.29   & 96     & 91.93   & 92.76   & 91.39   & 92.07  \\ \cline{2-10} 
                             & BT-GMod                 & 95.23   & 95.51   & 95.12   & 95.31  & 92      & 91.94   & 91.56   & 91.75  \\ \hline
\multirow{2}{*}{Both}        & BT-PG                   & 95.53   & 96.81   & 95.35   & 96.07  & 92.49   & 93.4    & 92      & 92.7   \\ \cline{2-10} 
                             & BT-PG-EG                & 95.41   & 97.07   & 95.13   & 96.09  & 92.45   & 93.73   & 91.86   & 92.79  \\ \hline
\end{tabular}%
}
\caption{Models' results on \emph{SError} set on character-level metrics.}
\label{sec}
\end{table*}

\section{Appendix: SIGHAN Case Study}
\label{sec:sighan}
We conduct a random sampling of the target sentences in the SIGHAN2015 test set and list the errors we find in Table~\ref{sighan error}.

\begin{table*}[]
\centering
\resizebox{\textwidth}{!}{%
\begin{tabular}{cc|c}
\hline
\multicolumn{1}{c|}{PID}                             & Corrected Sentence in SIGHAN2015   Test set                                                                                                                & Error                                                                                                                                                 \\ \hline
\multicolumn{1}{c|}{A2-0092-2}   & 他戴著眼镜{\color[HTML]{FE0000}跟袜子}入睡了。                                                                                                                                               & Grammar Error                                                                                                                                         \\ \hline
\multicolumn{1}{c|}{Explanation} & He went to sleep wearing his glasses and socks.                                                                                                            & \begin{tabular}[c]{@{}c@{}}The socks can't be 戴(worn) in Chinese, \\ but should be 穿(worn).\end{tabular}                                              \\ \hline
\multicolumn{1}{c|}{A2-1054-1}   & 我喜欢{\color[HTML]{FE0000}飞机}台湾。                                                                                                                                                   & Grammar Error                                                                                                                                         \\ \hline
\multicolumn{1}{c|}{Explanation} & I like to plane Taiwan.                                                                                                                                    & It should be  我喜欢飞到(fly   to)台湾.                                                                                                                      \\ \hline
\multicolumn{1}{c|}{B2-1934-2}   & 他可能因为意识到钱不见而心{\color[HTML]{FE0000}理}方寸大乱。                                                                                                                                        & Spelling Error                                                                                                                                        \\ \hline
\multicolumn{1}{c|}{Explanation} & \begin{tabular}[c]{@{}c@{}}He may have been disoriented by the realization\\ that the money   was not there.\end{tabular}                                  & \begin{tabular}[c]{@{}c@{}}心理(Psychology) should be \\ changed to  心里(in heart).\end{tabular}                                                         \\ \hline
\multicolumn{1}{c|}{B2-2241-1}   & \begin{tabular}[c]{@{}c@{}}经过{\color[HTML]{FE0000}只}么多苦，他们在大学\\ 有比较好的教育。\end{tabular}                                                                                            & Spelling Error                                                                                                                                        \\ \hline
\multicolumn{1}{c|}{Explanation} & \begin{tabular}[c]{@{}c@{}}After so much suffering,\\ they have a better education at the   university\end{tabular}                                        & 只(only) should be changed to 这(so).                                                                                                                   \\ \hline
\multicolumn{1}{c|}{B2-3835-3}   & \begin{tabular}[c]{@{}c@{}}我可以轻松地跟家人{\color[HTML]{FE0000}连}络，\\ {\color[HTML]{FE0000}网路}的资讯对我功课帮助很大。\end{tabular}                                                                                      & Language Conventions and Spelling Error                                                                                                               \\ \hline
\multicolumn{1}{c|}{Explanation} & \begin{tabular}[c]{@{}c@{}}I can easily communicate with my family \\ and the information on the Internet \\ helps me a lot with my homework.\end{tabular} & \begin{tabular}[c]{@{}c@{}}连(link) should be changed \\ to 联(communicate). \\ 网路 always are written as 网络(Web) \\ in Simplified Chinese.\end{tabular} \\ \hline
\multicolumn{1}{c|}{B2-3848-1}   & \begin{tabular}[c]{@{}c@{}}在我的国家{\color[HTML]{FE0000}也}电脑\\ 网路是青少年的生活中最重要的{\color[HTML]{FE0000}品}。\end{tabular}                                                                                        & Grammar Error and Language Conventions                                                                                                                \\ \hline
\multicolumn{1}{c|}{Explanation} & \begin{tabular}[c]{@{}c@{}}In my country also computers and \\ the Internet are the most important items \\ in the life of young people.\end{tabular}      & \begin{tabular}[c]{@{}c@{}}It should be 我的国家电脑网路\\ 也是青少年的生活中最重要的物品.\end{tabular}                                                                      \\ \hline
\multicolumn{1}{c|}{B2-4149-3}   & 这{\color[HTML]{FE0000}两}问题真的严重，我们受不了。                                                                                                                                            & Spelling Error                                                                                                                                        \\ \hline
\multicolumn{1}{c|}{Explanation} & \begin{tabular}[c]{@{}c@{}}These two problems are really serious\\ and we can't stand them.\end{tabular}                                                   & \begin{tabular}[c]{@{}c@{}}两(two) should be changed \\ to 俩(two) or 两个(two).\end{tabular}                                                             \\ \hline
\multicolumn{1}{c|}{B2-4265-1}   & 孩子会一直依赖{\color[HTML]{FE0000}著}父母过生活。                                                                                                                                            & Language Conventions                                                                                                                                  \\ \hline
\multicolumn{1}{c|}{Explanation} & \begin{tabular}[c]{@{}c@{}}The child will always be dependent\\ on the parent to live his or her life.\end{tabular}                                        & \begin{tabular}[c]{@{}c@{}}著 always are written as 着 \\ in Simplified Chinese.\end{tabular}                                                           \\ \hline
\end{tabular}%
}
\caption{Error examples in the target sentences in SIGHAN2015 test set. Some sentences are truncated due to length, and only the problematic fragments are shown, which do not affect the semantics.}
\label{sighan error}
\end{table*}

\section{Appendix: BART Result on SIGHAN}
To verify the cheat-like effect of the dataset Wang271K on SIGHAN, we conducted an experiment with BART. The results are shown in Table~\ref{tab:bartres}.
% Please add the following required packages to your document preamble:
% \usepackage{graphicx}
\begin{table*}[]
\centering
\resizebox{\textwidth}{!}{%
\begin{tabular}{cc|cccc|cccc}
\hline
Training   Set                       & Test Set  & \multicolumn{4}{c|}{Detection Level} & \multicolumn{4}{c}{Correction Level}       \\ \hline
\multicolumn{1}{c|}{}                &           & Acc.    & Pre.    & Rec.    & F1     & Acc.   & Pre.   & Rec.   & F1              \\ \hline
\multicolumn{1}{c|}{SIGHAN}          & SIGHAN13  & 0.384   & 1       & 0.384   & 0.5549 & 0.282  & 1      & 0.282  & 0.4399          \\ \hline
\multicolumn{1}{c|}{SIGHAN}          & SIGHAN14  & 0.6902  & 0.7837  & 0.5254  & 0.6291 & 0.6629 & 0.7645 & 0.4708 & 0.5828          \\ \hline
\multicolumn{1}{c|}{SIGHAN}          & SIGHAN15  & 0.7555  & 0.863   & 0.6073  & 0.7129 & 0.6936 & 0.8339 & 0.4836 & 0.6122          \\ \hline
\multicolumn{1}{c|}{SIGHAN+Wang271K} & SIGHAN13  & 0.571   & 1       & 0.571   & 0.7269 & 0.556  & 1      & 0.556  & \textbf{0.7147} \\ \hline
\multicolumn{1}{c|}{SIGHAN+Wang271K} & SIGHAN14  & 0.7354  & 0.799   & 0.629   & 0.7039 & 0.7269 & 0.7946 & 0.6121 & \textbf{0.6915} \\ \hline
\multicolumn{1}{c|}{SIGHAN+Wang271K} & SIGHAN15  & 0.8073  & 0.8521  & 0.7436  & 0.7942 & 0.7909 & 0.8463 & 0.7109 & \textbf{0.7727} \\ \hline
\multicolumn{1}{c|}{SIGHAN+Wang271K} & Our\_test & 0.1933  & 1       & 0.1933  & 0.324  & 0.1765 & 1      & 0.1765 & 0.3001          \\ \hline
\end{tabular}%
}
\caption{BART results on SIGHAN and our test set. The training set is SIGHAN training set and Wang271K data.}
\label{tab:bartres}
\end{table*}
\label{sec: bart}

\end{CJK*}
\end{document}